\documentclass[11pt]{article}

\usepackage[final]{acl}

\usepackage{times}
\usepackage{latexsym}

\usepackage[T1]{fontenc}

\usepackage[utf8]{inputenc}

\usepackage{microtype}

\usepackage{inconsolata}

\usepackage{graphicx}
\usepackage{booktabs,tabularx, multirow}

\usepackage{paralist}
\usepackage{enumitem}

\usepackage{color-edits}
\addauthor{hh}{magenta}
\addauthor{hs}{purple}
\addauthor{cs}{cyan}
\addauthor{ml}{blue}
\addauthor{nt}{orange}

\title{Between Rigor and Reality: How AI Safety Benchmark Developers Understand Benchmark Use and Maintenance}

\author{Megan Li\thanks{\ \ Co-first authors; equal contribution}, Ningjing Tang\footnotemark[1], Christopher Stewart\footnotemark[1], \\ \textbf{Hoda Heidari\thanks{\ \ Co-last authors; equal contribution}, \and Hong Shen\footnotemark[2]} \\
        Carnegie Mellon University\thanks{\ \ Presented as a poster at the Eval Eval Workshop at ACL 2026.} \\  Pittsburgh, PA \\
\small{
   \textbf{Correspondence:} \href{mailto:meganli@andrew.cmu.edu}{meganli@andrew.cmu.edu}, \href{mailto:ningjint@andrew.cmu.edu}{ningjint@andrew.cmu.edu},
   \href{mailto:cstewar3@andrew.cmu.edu}{cstewar3@andrew.cmu.edu}
 }
}

\begin{document}
\maketitle
\begin{abstract}
Safety benchmarks are cited as the primary tool to understand the risks posed by generative AI (genAI) systems, yet growing evidence suggests they often fail to meet the needs of real-world safety evaluation.  We present findings from five interviews with AI safety benchmark developers in academia and industry. We find a disconnect between perceptions of the \textit{intended} usefulness of safety benchmarks and their \textit{practical} usefulness in safety evaluations of deployed systems. Our participants offered perspectives on two challenges contributing to this disconnect: (1) the difficulty of achieving inter-rater reliability on safety constructs, and (2) a lack of clarity regarding how to address persistent threats to external validity. Based on our analysis, we argue that safety benchmarks must not only be grounded in deployment contexts, but \textit{actively integrated} with them. To facilitate this, we outline pathways for more transparent communication between academic and industry AI safety stakeholders.
\end{abstract}

\section{Introduction}
As generative AI (genAI) systems are adopted across diverse contexts, characterizing the safety of these systems becomes increasingly relevant \cite{safetybench,weidinger_sociotechnical}. 
Prior work defines AI safety as the endeavor to prevent or mitigate harms from AI systems \cite{harding2025whatisaisafety}; researchers in this field have developed a range of approaches toward this end, including the development and use of safety benchmarks.
Safety benchmarks combine curated datasets and scoring metrics to evaluate model behavior in risk-relevant scenarios \cite{everything_benchmark, more_than_marketing}. They have been developed to cover diverse dimensions of risk, including toxicity, bias, robustness, and harmfulness \cite{safetybench,MLcommons_safety,harmbench}.

While safety benchmarks are cited as the most common solution to evaluate risks posed by AI systems, there is increasing evidence that they often fail to meet real-world needs, especially in safety-critical contexts \cite{more_than_marketing,howshouldaibenchmark}.
For example, recent work has found that AI safety benchmarks focus on known risks to the exclusion of less predictable harms, reduce complex scenarios to simplistic pass/fail decisions, and fail to provide meaningful context for the scores they produce \cite{howshouldaibenchmark,canwetrustaibenchmarks}.
\citet{more_than_marketing} found that practitioners who use (or decide not to use) benchmarks in their day-to-day work are generally aware of these limitations, often leading them to consider benchmarks insufficient for informing decisions. 

However, we lack research documenting how benchmark \textit{developers}, who are uniquely positioned to shed light on the incentives behind and practical realities of developing these benchmarks, view these issues. 
Because benchmarks sit at the intersection of academic methodology and industry practice \cite{aiasasport}, benchmark developers must navigate diverse and potentially competing
disciplinary norms, organizational constraints, and assumptions about how their benchmarks will be used. Their perspectives are essential to understand \textit{why} benchmarks exhibit the limitations identified in previous work and provide recommendations that are grounded in their practical constraints.
This is the focus of our work, framed by two research questions:

\begin{itemize}
    \item \textbf{RQ1:} What needs do benchmark developers perceive safety benchmarks are \textit{intended} to meet? How do they compare to the needs they perceive safety benchmarks to meet \textit{in practice}?
    \item \textbf{RQ2:} How do they currently interpret and/or address the challenges that contribute to gaps between safety benchmarks and real-world needs?
\end{itemize}


To answer these questions, we conducted five semi-structured interviews with AI safety benchmark developers working in academia and industry. Our findings reveal a tension between how benchmark developers perceive the intended usefulness of safety benchmarks and how they perceive their practical usefulness in real safety evaluations for deployed AI systems. 
We offer benchmark developers' perspectives on two factors contributing to this tension: first is the challenge of achieving satisfactory inter-rater reliability in benchmark development, and second is a lack of clarity regarding how to address persistent threats to the external validity of safety benchmarks.
We present this work as an early snapshot of 
the gap between benchmarks' perceived authority as measures of safety and their seeming incompatibility with the dynamic, context-dependent evaluation practices that are relied upon to assess deployed genAI systems.

\section{Related Work}
\subsection{Challenges of developing and maintaining safety benchmarks}
\label{challenges-relatedwork}

There are many known shortcomings of safety benchmarks, including poor construct validity and becoming quickly outdated \citep{safetywashing, canwetrustaibenchmarks, howshouldaibenchmark, hatevolution, xu2024benchmarkingbenchmarkleakagelarge, Fodor2025-cn}. Throughout this work, we frame these as the result of challenges that benchmark developers grapple with in the \textit{development} and \textit{maintenance} of their benchmarks. As we discuss in this subsection, not only do safety benchmarks inherit the challenges of AI benchmarks in general---for instance, issues of data leakage and saturation render many publicly available benchmarks invalid \citep{xu2024benchmarkingbenchmarkleakagelarge, Fodor2025-cn, canwetrustaibenchmarks}---but they also bring unique challenges due to the context-dependent nature of safety constructs.

\subsubsection{Challenges arising during benchmark development}

A growing body of work has highlighted several challenges that arise during benchmark development, including establishing construct validity (i.e., ``the extent to which what was to be measured was actually measured'' \cite{validity_assessment, def_construct}) and achieving satisfactory inter-rater reliability (i.e., the degree to which a measure is sensitive to properties of the specific annotators who labelled data \cite{reliability}) \citep{howshouldaibenchmark, safetywashing, canwetrustaibenchmarks, aisafetycascading}. For example, concerns have been raised about whether benchmarks can meaningfully capture the complexity of safety-relevant phenomena, including whether high scores reflect genuine safety or merely surface-level compliance \citep{safetywashing, canwetrustaibenchmarks}. One of the reasons for these concerns is that the concepts AI safety benchmarks attempt to measure tend to be more subjective than those addressed by other kinds of benchmarks. For example, \citet{howshouldaibenchmark} examined 210 safety benchmarks and found that 94\% disclosed uncertainty about the labels of benchmark items, often using metrics like inter-rater reliability minima to exclude borderline cases. Unlike capability benchmarks, where correctness may be more readily defined, safety benchmarks are evaluating inherently normative, context-dependent constructs, making the challenge of achieving satisfactory reliability especially demanding.

\subsubsection{Challenges arising after benchmark publication and/or deployment}

AI safety benchmarks often measure concepts that shift as societal contexts change and use cases evolve. As a result, benchmarks that were once meaningful proxies for safety may become increasingly misaligned with the harms they purport to measure \citep{hatevolution, weidinger_sociotechnical}. For example, \citet{hatevolution} demonstrate that static benchmarks are poorly suited to capturing rapidly evolving socio-cultural phenomena such as hate speech, whose forms, targets, and community-specific meanings shift over time in ways that fixed datasets cannot track. Their work illustrates how a benchmark that was valid at one point in time may yield misleading assessments as linguistic norms and harmful behaviors evolve, a process referred to as ``concept shift.'' Even labels that remain conceptually valid may fail to reflect the actual inputs that a model encounters in practice. As models are deployed across platforms, languages, and user populations not represented in the original dataset, real-world inputs increasingly diverge from the benchmark's test distribution \cite{zhang2023dive}.

Beyond concept shift and distribution shift, benchmarks face threats from data leakage and saturation \citep{xu2024benchmarkingbenchmarkleakagelarge, Fodor2025-cn, canwetrustaibenchmarks}. When benchmark datasets become publicly available, they risk being absorbed into model training corpora, rendering them ineffective as held-out evaluation sets. Research has shown that saturation similarly undermines a benchmark's ability to discriminate between systems or measure further progress \citep{benchmarksaturation, canwetrustaibenchmarks}. However, it is unclear how benchmark developers view these issues and how it affects AI safety evaluation more broadly.

Despite growing recognition of these threats, the field lacks established norms for when and how benchmarks should be retired, revised, or supplemented. Benchmarks continue to be cited and their datasets reused long after the conditions under which they were validated have changed, with no systematic mechanism for signaling when a benchmark's useful life has expired. Unlike software libraries, which benefit from versioning conventions and deprecation notices, benchmarks lack analogous infrastructure for communicating their current fitness for use \citep{cogo2021deprecation}. This context creates threats to external validity. Our paper takes this gap as a starting point, asking how practitioners themselves understand and navigate the temporal limitations of the benchmarks they build and maintain.

\subsection{Gaps in our understanding of safety benchmark practices}

The shortcomings documented above have largely been identified through surveys of benchmark artifacts \citep{howshouldaibenchmark, canwetrustaibenchmarks} or through interviews with those who use (or choose not to use) benchmarks \citep{more_than_marketing}. Less is known about how the people who build safety benchmarks understand these limitations and navigate them in their own work. 

Emerging evidence suggests that evaluation practices in industry are diverging from the benchmarking paradigm. Practitioners often develop internal evaluation tools that they consider more relevant than public benchmarks, yet these tools are rarely shared or externally scrutinized \citep{more_than_marketing}. Some evidence suggests that safety evaluation in practice is often an \textit{ad hoc} process that shares more commonalities with the \textit{bricolage} paradigm of measurement and target variable construction used by data scientists \citep{bricolage} than the principled, top-down approach increasingly advocated by scholars \citep{wallach2025, weidinger_sociotechnical}. Understanding how benchmark creators themselves experience this tension---between the standards their benchmarks are meant to uphold and the realities of evaluation in practice---is the central aim of this work.
\section{Methods}
\label{methods}

We conducted five semi-structured interviews with practitioners who reported having experience developing or maintaining safety benchmarks---defined in our recruitment call as ``a standardized test or evaluation framework to measure a system's propensity to exhibit unsafe behavior, such as by producing misinformation or hate speech''---for GenAI. We recruited participants through our professional networks and social media. All interviews were approximately an hour long and conducted virtually over Zoom between November 2025 and January 2026. Participants provided informed consent before their
interviews and we compensated each participant for their time with a \$60 gift card. The study was approved by our institution’s IRB.

\paragraph{Participants} We summarize our participants' professional contexts and benchmark focuses in Table~\ref{tab:participants-summary}. Participants had experience developing benchmarks in industry and academic contexts to measure concepts related to mental health, bias, and toxicity.

\begin{table}[h!]
    \centering
    \begin{tabular}{p{0.1\linewidth}p{0.25\linewidth}p{0.3\linewidth}p{0.15\linewidth}}
    \toprule
         \textbf{ID} & \textbf{Professional Context} & \textbf{Focus} & \textbf{YoE} \\\midrule
         P1 & Industry & Evaluation methodology & 10+ \\
         P2 & Academia & Mental health & 1-5 \\
         P3 & Industry* & Bias, Toxicity & 1-5 \\
         P4 & Industry & Human well-being & 1-5 \\
         P5 & Industry* & Pluralism & 5-10 \\ \bottomrule
    \end{tabular}
    \caption{Participants' professional contexts, benchmark focuses, and years of experience developing evaluations (YoE). P4 and P5 had significant experience in both academic and industry contexts, so we indicate their context at the time of their interview.}
    \label{tab:participants-summary}
\end{table}

\paragraph{Interview protocol} 
Following established qualitative research methods for semi-structured interviews, we designed our protocol to begin with broad, open-ended questions about participants’ backgrounds and current practices with GenAI benchmarks, examining the benchmarks they had developed or regularly used and how these fit into their development workflows. To establish a shared frame of reference, we provided a working definition of a GenAI benchmark as ``a standardized evaluation framework consisting of a dataset of prompts or tasks and a metric for scoring model performance.'' We maintained flexibility to probe deeper through follow-up questions, particularly around dimensions identified in prior literature on sociotechnical evaluation, including benchmark validity, contextual adaptation, and stakeholder decision-making.

To investigate benchmark maintenance under real-world validity threats over time, we specifically designed questions that explored how practitioners assess whether a benchmark remains aligned with its intended construct over time, what signals trigger updates, and how update decisions are made and by whom. We explicitly scoped the discussion to concept drift (the evolution of real-world meaning) rather than benchmark contamination from training data leakage. For example, when participants described past benchmark updates, we probed what triggered the decision, who was involved, what evidence informed the revision, and how its effectiveness was evaluated.

For participants who had not encountered concept drift directly, we introduced one of two hypothetical scenarios to elicit practical reasoning. For each, we probed how teams would detect the gap, what criteria would govern the update decision, and what downstream consequences should be anticipated. The protocol concluded by asking participants to reflect on the risks of inadequate benchmark maintenance and to envision ideal systems for keeping benchmarks current, including the appropriate role of automation and human oversight.

\paragraph{Analysis} We analyzed our interview transcripts using reflexive thematic analysis \cite{Braun08082019}. The three first authors performed an initial open coding of all the interview transcripts and kept detailed memos throughout the coding process to document emerging patterns and potential themes. The codebook was developed iteratively by the three co-first authors through this open coding process, beginning with descriptive codes grounded in the data and progressively developing more analytical codes. The research group also held regular collaborative sessions to discuss emerging codes, resolve discrepancies and ambiguities in the codes, and iteratively refine our coding scheme. 
\section{Findings}


Our findings suggest an increasing divergence between the needs that participants imagine safety benchmarks are intended to meet and the needs they perceive these benchmarks to meet in practice. We highlight how participants interpret and address two challenges that arise in benchmark development: (1) the difficulty of achieving inter-rater reliability when creating safety benchmark datasets, and (2) persistent threats to the external validity of safety benchmarks; as well as how these challenges contribute to the gap between intended and practical uses of safety benchmarks.

\subsection{Safety benchmarks: intention versus reality} \label{intended-use-reality}

Our participants discussed several intended and potential uses for safety benchmarks. Yet, participants reported lacking feedback on whether the benchmarks they published were actually being used in these ways, and when describing their own safety evaluation practices, our participants reported that they did not use published benchmarks. 

\paragraph{Intended and potential uses of safety benchmarks include measuring and guiding progress, comparing models, and holding developers accountable.} Participants discussed that in an ideal world, intended uses of safety benchmarks include hill-climbing, measuring progress, and comparing across industry. This largely reflects how ML and computer science communities at large conceptualize the intended uses of benchmarks \cite{benchmarking_cs_history, everything_benchmark, CHURCH_2018}. 

Distinct from previous literature on intended uses and norms of benchmarks, 
P5 thought that \textit{safety} benchmarks specifically could be useful \textit{``to hold...model developers...accountable in a way that is not influenced by their own model development.''} P5 also proposed speculative uses, including to operationalize government regulations and incorporate participatory methods into model evaluation, thus steeping evaluation in the realities of regulation and users. 

\paragraph{Participants often lacked feedback as to how their safety benchmarks were being used.} Despite articulating several intended uses for safety benchmarks, participants were unsure whether the benchmarks they had developed were being used in these ways or whether they were being used at all. This was true even for participants who described experiences tailoring or piloting their benchmarks with real clients (P3 and P4), suggesting that participants desired broader, community- or industry-level adoption and feedback on how their work was being used. For instance, P3 shared that \textit{``We didn't really know how our benchmark was being used, or if it was being adopted. We put it on HuggingFace and we presented at a conference and talked to people, but we weren't really sure if it was being used by other model developers or startups or other groups.''} This perspective has not been documented in previous work on AI benchmarks.

\paragraph{Some participants expressed skepticism about the meaningfulness of safety benchmarks.} P1 explained that it is generally unclear what the outcome of safety benchmark means---that is, there are no clear consequences for receiving a `bad' versus `good' score on such a benchmark---except for allowing a corporation or company to say \textit{``oh, my product is okay.''} He perceived that safety benchmarks are often developed and publicized by corporations \textit{``as a part of PR campaign.''} P5 shared a similar concern that safety benchmarks could give model developers \textit{``a false sense''} of safety. However, while P1's concern was rooted in a perception about incentives behind safety benchmark development, P5 explained that her concern was due to having often found safety benchmarks to be \textit{``quite lacking.''} Both perspectives reflect skepticism about the meaningfulness, utility, and rigor of published safety benchmarks. 

\paragraph{Industry practitioners conduct safety evaluations that do not make use of benchmarks.} 

Our participants in industry contexts emphasized that safety evaluations serve real company needs: in particular, they are necessary to inform consequential decision-making. The evaluations that they highlighted to serve these needs, however, do not include published benchmarks. Instead, our participants described maintaining internal evaluation ecosystems that make use of red-teaming, user logs, and user feedback, which they positioned in contrast to \textit{``[static benchmarks] which people publish in journals or conferences''} (P5).



Our participants' experiences may in part be explained by the well-documented divergence between standardized, public-facing assessment and more pragmatic, internally-oriented testing that has emerged in computing since the 1980s \cite{benchmarking_cs_history, more_than_marketing}. However, 
our findings suggest that this divergence is especially pronounced in AI safety, \textbf{where it is increasingly unclear whether these divergent approaches to safety evaluation complement, contradict, or obviate each other.} In the next two subsections, we characterize two challenges that arise in the development of a safety benchmark and how they may contribute to this divergence.


\subsection{Achieving and interpreting inter-rater reliability in safety benchmarks}
Participants highlighted inter-rater reliability as a persistent challenge when developing safety benchmarks. Participants interpreted this challenge in two distinct ways: as a \textit{bug}---that is, a barrier to overcome in the endeavor to establish a valid benchmark---or as a \textit{feature}---that is, a signal that indicates inherent pluralism in how human raters engage with the measured concept. 

\paragraph{Participants highlighted inter-rater reliability as a persistent challenge in the development of safety benchmarks.}
P2 illustrated the challenge of reaching inter-rater reliability through his work in the mental health domain, noting that even domain experts struggle to reach consensus on what constitutes harm: \textit{``even within the...clinical psychiatrist community, they also struggled to identify harm just within that domain.''} He described the extensive efforts required to achieve satisfactory inter-rater reliability on the expert annotations collected for his benchmark: \textit{``We actually conducted a lot of training sessions [with the expert annotators] where we have big group discussion...Everybody can voice their opinion. We talk a couple of times, which is extremely hard to organize, given everybody's availability...And also we have multiple rounds of annotation too.''}
P2's experience indicates that unlike capability benchmarks, where correctness may be more readily defined, safety benchmarks are evaluating inherently normative, context-dependent constructs, making the challenge of achieving satisfactory reliability extremely demanding.

\paragraph{Some participants viewed the difficulty of achieving inter-rater reliability as a barrier to overcome.}
P1 crystallized this idea when discussing barriers to developing valid benchmarks, noting that \textit{``validity is an extremely high bar, and I was happy in my benchmarking career to see that people had made attempts to address \textbf{reliability, which is a necessary but not sufficient condition for validity.}''} He discussed finding that even some of the most-cited benchmarks where the ground truth is readily accessible (e.g., for image classification) suffered from significant inter-rater reliability problems, and that this problem is even larger in \textit{``more subjective''} contexts such as safety. He expressed a desire for the field to acknowledge and dedicate efforts toward solving this problem. P2's efforts to achieve satisfactory inter-rater reliability reflect this sentiment, conceptualizing this type of reliability as a prerequisite for a useful benchmark.

\paragraph{Others viewed the difficulty of achieving inter-rater reliability as a reflection of the inherent pluralism of safety.} Participants generally agreed that safety evaluations often involved measuring contested or subjective concepts, on which human opinions are known to be pluralistic. For some participants, this meant viewing low agreement between human raters as a feature of the task rather than a bug to be fixed. For instance, P5 noted, \textit{``safety is definitely a subjective thing,''} and grounded this insight in the following example: \textit{``the kinds of stereotypes that one might consider in a western context are not maybe the same for people in different parts of the world.''} Rather than thinking of this pluralism as a barrier to overcome, P5 indicated that she felt benchmarks should therefore capture a wider range of perspectives in order to be useful. That is, an ideal safety benchmark would have \textit{``exhaustive coverage''} of the distribution of possible perspectives rather than attempting to enforce agreement to establish a `correct' perspective.





\subsection{Acknowledging threats to the external validity of safety benchmarks}

Participants discussed several potential threats to the external validity of a safety benchmark. They acknowledged that, especially in the domain of safety, distribution shift and concept shift posed threats to the external validity of a benchmark. However, they raised data leakage, Goodhart's Law, and saturation---all well-documented threats to the validity of benchmarks in general (see subsection~\ref{challenges-relatedwork})---as more immediate concerns when considering external validity. Yet, these concerns did not tend to motivate participants to actively maintain the external validity of their benchmarks (i.e., by updating them over time). 

\paragraph{In the context of safety, participants acknowledged that distribution shift and concept shift can degrade a benchmark's validity.} P1 and P2 each described particular examples of distribution shift that they had either encountered in their own work or observed online; P1 described how the recent prevalence of \textit{``bad advice on on...drugs like Ozempic and Wegovy''} represented a shift in the distribution of harmful online content, and P2 described how his work felt like \textit{``playing catch up with [online] communities...they're going to come up with new methods to get around safeguards [that identify harmful content].''} 
P1, P3, and P5 noted that concept shift was an expected phenomenon in the domain of safety and noted that such shift would occur not just over time but across contexts. P5 further noted that concept shift could sometimes be a direct result of a model's diffusion in the world, positing that people \textit{``have different expectations from [models]...depending on the model's stay or amount of use in that community,''} resulting in shift in the concept of ``safety'' and how it applies to a model or type of model over time. 





\paragraph{Participants raised data leakage, Goodhart's Law, and saturation as key concerns when considering the relevance and usefulness of benchmarks.} We remark that this was despite our attempts to scope the discussion around concept drift (see Section~\ref{methods}), likely because our participants encountered these concerns more concretely and more frequently than concept drift. 

P1 asserted that due to current training data acquisition practices, the problem of data leakage is so prevalent and immediate that \textit{``the only moment in which [a benchmark prompt] is valid is when you have not shared that prompt with anyone''} since publicized prompts \textit{``will immediately be captured by the [models] and the [models] won't trip up on those particular examples.''} P1 explained that this led him to estimate a corresponding \textit{``expiration date''} for each benchmark dataset he published, thus conceptualizing benchmark development as a recurring rather than one-time cost. P5 raised Goodhart's Law as a reason that benchmarks lose their utility, arguing \textit{``anything you measure is a bad thing to measure after you've measured it enough times.''} P1, P3, and P5 described saturation as a common reason for benchmark deprecation, as it rendered benchmarks useless for both hill-climbing and comparison (since, as P1 stated, \textit{``all the models converge on performing very, very well''}). These concerns largely echo those documented in previous literature on the limitations of ML benchmarks in general (see subsection~\ref{challenges-relatedwork}).

\paragraph{These concerns did not motivate participants to actively maintain their benchmarks' external validity.} Despite participants' awareness and acknowledgement of persistent threats to safety benchmarks' external valdiity, only P1 reported responding to these threats by regularly updating his benchmark. All other participants reported that they had not updated their benchmarks since publication and had no plans to do so. 

Overall, participants indicated that this was due to insufficient incentives and/or exorbitant time and resource costs associated. In particular, P3 explained that while he had initially planned to improve and maintain his benchmark over time, a lack of feedback as to whether or how his benchmark was being adopted (see subsection~\ref{intended-use-reality}) led to a lack of incentive to do so. P2 noted that new, unseen data was necessary to inform such updates, and that in his setting, such data was difficult to obtain.

This is in sharp contrast to the safety evaluations that our participants described as common in industry contexts. These evaluations were characterized as \textit{``a developing ecosystem which constantly adapts, adapt, adapts, adapts''} (P5) and, as described in subsection~\ref{intended-use-reality}, may incorporate several methods (red-teaming, monitoring user logs, monitoring activity responses or feedback) to inform this adaptation. This contrast suggests that for benchmark developers, a lack of adoption and feedback (as raised by our participants in subsection~\ref{intended-use-reality}) directly undermines the incentives and resources needed for benchmark maintenance. More broadly, it gestures to the idea that safety benchmarks can best meet real-world needs when they are embedded in evaluation infrastructures for deployed systems, where relevant safety constructs, gaps in coverage, and associated consequences are made concrete. 
\section{Discussion}
Our findings reveal a tension in the field of AI safety evaluation: safety benchmarks are imagined as tools for measuring progress, enabling comparability, and holding developers accountable, yet our participants---themselves safety benchmark developers---neither use them when evaluating deployed systems nor maintain them over time. Drawing on their experiences developing these benchmarks, our participants shed light on how this tension arises and why it persists. We discuss two implications of our findings for how the field conceptualizes AI safety evaluation.



\subsection{The reality of safety evaluation may necessitate a bricolage approach}

Our findings suggest that the limitations of public safety benchmarks stem not only from the difficulty of measuring safety, but from their disconnection from the real-world feedback loops that keep evaluation relevant. Our participants were aware of persistent threats to their benchmarks' validity, yet lacked the signals that would indicate to them when their benchmarks had drifted from practice or fallen out of use altogether.
Against a backdrop of calls from academia to develop a rigorous, top-down approach to evaluating generative systems \cite{wallach2025, evaluation_science}, we argue that this disconnect warrants interrogation. While top-down approaches offer methodological rigor and increased transparency, our findings suggest that they are inflexible to post-deployment feedback. Thus, we instead propose a \textit{bricolage} mode of safety evaluation that may begin with a rigorous theory, but is iteratively refined from the bottom-up in concrete deployment contexts \cite{bricolage}.

As our findings on the challenge of reaching inter-rater reliability illustrate, safety's inherently normative and context-dependent character makes top-down standardization particularly difficult to sustain. Standardized benchmarks implicitly require a shared understanding of harm often without reference to the concrete use contexts in which a model will be deployed \cite{Narayanan2024-rl}, yet the disagreements between annotators may reflect genuine different views of safety between contexts and communities \cite{pluriharm}. A top-down benchmark cannot easily accommodate this pluralism, whereas internal evaluations grounded in specific deployment contexts can be calibrated to the norms and expectations of the populations they serve.

A key mechanism sustaining valid safety evaluation in this bricolage model is continuous feedback from deployment contexts. Our findings complement \citet{more_than_marketing}, who demonstrated from the benchmark \textit{user} side that practitioners supplement or abandon public benchmarks in favor of internal evaluations. We show that this disengagement extends to the developers themselves, driven by an absence of feedback from deployment contexts that leaves developers with neither the incentive to maintain their benchmarks nor the signals needed to know whether they remain valid. Building on calls for GenAI safety evaluations to consider deployment contexts \cite{Narayanan2024-rl, evaluation_science}, we argue that it is necessary for safety benchmarks to be \textit{actively integrated} in deployment contexts, with explicit mechanisms soliciting signals over time that can inform when a benchmark needs revision or retirement.

This does not mean abandoning structured, public-facing benchmarks. Rather, we suggest the field needs to reconceptualize academic benchmarks and internal evaluations as complementary components of an evaluation ecosystem.





\subsection{A need for transparent communication between industry and academia}
Our findings raise concerns about the opacity of industry safety evaluation practices. While industry participants described heavy reliance on internal evaluation ecosystems, these systems are rarely shared or externally scrutinized. 
This opacity impedes the field's collective ability to understand what safety evaluation looks like in practice. \citet{the_walled_garden} provides a useful frame for understanding the challenges of this opacity, noting the structural barriers that make research on AI practice difficult when researchers wish to engage.

We call for greater transparency from industry on safety evaluation practices, which can take several forms:

\paragraph{Methodological transparency on internal safety evaluations.} Our findings suggest that internal safety evaluations often adopt signals like user logs and red-teaming findings for continuous updates, yet these practices are rarely documented or shared externally. We call on industry practitioners to publish high-level descriptions of their safety evaluation methodology. Model cards already provide vehicles for such disclosure \cite{modelcards}, but stronger norms are needed around reporting concrete safety evaluation practices from industry.

\paragraph{Closing the feedback loop of benchmark development.} Our findings show that benchmark developers frequently lack signal as to whether or how their benchmarks are being adopted, which results in little incentive to maintain them. A realistic channel for closing this feedback loop is third-party institutional involvement. AI governance bodies and standards organizations such as the National Institute of Standards and Technology (NIST), which already engage both academic researchers and industry practitioners in the development of evaluation frameworks, are well-positioned to collect and disseminate this kind of feedback systematically. Recent efforts by NIST \cite{NIST_AI_800-2_ipd} suggest this momentum is building, and greater institutional engagement of this kind could meaningfully strengthen the incentive structures that sustain benchmark ecosystems over time.

\paragraph{Collaborative standard sharing on benchmark updates and retirement.} AI safety---and AI evaluation more broadly---currently lacks established norms for when and how benchmarks should be retired, revised, or supplemented. We call for structured academic–industry collaborations to develop shared standards that define the conditions under which a benchmark should be updated or retired. For instance, these conditions could include evidence of saturation, concept shift, or systematic divergence between benchmark distributions and real-world inputs. 


\subsection{Limitations}

Our work has several limitations. First, we were limited to a small number of participants. This was due in part to time constraints and the relative difficulty of securing the participation of industry practitioners, which is a documented challenge of work in this area \cite{the_walled_garden}. 
Our sample was also drawn mainly from our professional network and responses to social media posts, imposing a selection bias. Finally, as with other work that relies on self-reporting methods, our work is subject to potential bias due to participants' desire to appear competent \cite{social_desirability}. These limitations should be kept in mind  when drawing conclusions from this work. Future work may address these biases by recruiting more participants, including via academic-industry partnerships, as well as incorporating additional data collection methods to strengthen the claims made here.


\bibliography{custom}

\appendix

\section{Interview Questions}


\subsection{Introductory items}
\begin{enumerate}
    \item Consent and recording permission
    \item Participant background: role, company type, experience with ML/AI benchmarks
    \item We define a GenAI benchmark as a standardized test or evaluation framework used to measure and compare the performance, capabilities, and quality of GenAI models. It usually consists of a dataset (the collection of tasks or prompts used for evaluation) and a metric (the scoring system that measures how well the model performs on those tasks).
\end{enumerate}

\subsection{Exploration of Participant's Current Practice}

\textbf{Primary Question}
\begin{enumerate}[resume]
    \item Tell me about your experience and experience as they relate to GenAI benchmarks.
\end{enumerate}

\textbf{Follow-up prompts:}
\begin{enumerate}[resume]
    \item What benchmarks have you or your team contributed to developing?
    \item What benchmarks do you or your team regularly use?
    \item How do these benchmarks fit into your development process?
\end{enumerate}

\subsection{Validity and Monitoring }
\paragraph{Introduce the concept:} When we talk about benchmark ``validity,'' we mean how well it measures what it's supposed to measure in the current context. We’re specifically concerned with situations of concept shift where the real-world meaning of the target concept evolves over time. We are \textit{not} focusing on issues of data contamination.

\textbf{Questions:}
\begin{enumerate}[resume]
    \item Do you or your team care if a benchmark is still aligned with the initial concept it was developed to measure?
    \item How do you assess if a benchmark is still aligned with the initial concept it was developed to measure?
    \item Can you think of  signals or metrics that might indicate a benchmark needs updates or modifications?
    \item Have you encountered situations where a benchmark led to incorrect or misleading results due to changes in the real world context of use (for example, due to a recent event in the news)?
\end{enumerate}

\paragraph{If participant mentions updating/changing benchmarks:}
\begin{enumerate}[resume]
    \item Can you walk me through a specific example of when you updated or changed a benchmark? What triggered that decision?
    \item Who was involved in making the decision to modify the benchmark? How did they inform the decision to change the benchmark?
    \item Did you use specific metrics to determine if a modification was necessary? For instance, metrics could be supplied from online or offline experiments involving the benchmark. (Online experiments means data coming from live production; offline experiments means data come from development sandbox environments or staging)
    \item What was the process of updating the benchmark?
    \item How did you assess whether the update was effective?
\end{enumerate}

\paragraph{If participant says they don't update benchmarks or never thought about it:}
That's helpful to know. Let me share a hypothetical scenario instead to get your thoughts on how these situations might arise in practice.
\paragraph{Healthcare Misinformation Scenario:} Imagine someone has a benchmark dataset for detecting health-related misinformation. Initially, it contained various types of false health claims annotated as such. Then, a new type of misinformation becomes prevalent - let's say false claims about acetaminophen use during pregnancy causing autism in children. This wasn't in their original benchmark dataset, but now it's spreading widely and rapidly online, and users of GenAI models are frequently inquiring about the veracity of this claim.

\paragraph{Alternative Scenario (Evolution of Eating Disorder-related Language Online):} Imagine you have a benchmark dataset for detecting eating disorder-related content on social media platforms. Your original dataset was trained to recognize explicit terms and phrases people used to discuss disordered eating behaviors. However, online communities have evolved new coded language and euphemisms to avoid detection - terms like 'spicy water' for appetite suppressants, or using emojis and abbreviations that weren't in your initial benchmark dataset. Your benchmark is now missing these newer ways people discuss eating disorders.

\textbf{Questions:}
\begin{enumerate}[resume]
    \item If you were to give advice to the team, how do you think they would become aware of this gap?
    \item What would be the process for deciding whether to update the benchmark?
    \item Who would make that decision?
    \item What constraints or considerations would/should influence that decision?
    \item How would/should their team go about the process of updating the benchmark?
\end{enumerate}

\subsection{Downstream Impacts}
\begin{enumerate}[resume]
    \item What are some of the downstream impacts you would expect to see if a benchmark is not maintained appropriately?
\end{enumerate}



\end{document}